%% file: main.tex
\documentclass[9pt,conference]{IEEEtran}
\IEEEoverridecommandlockouts
\usepackage{cite}
\usepackage{amsmath,amssymb,amsfonts}
\usepackage{algorithmic}
\usepackage{graphicx}
\usepackage{textcomp}
\usepackage{booktabs, multirow}
\usepackage{authblk}
\usepackage{array}
\usepackage{pifont}
\usepackage{xcolor,colortbl}[table]
\usepackage[percent]{overpic}
\usepackage{subfig}
\usepackage{comment}
\usepackage{bbm, dsfont}

\usepackage{tablefootnote}
\usepackage{multicol}
\newcommand\blfootnote[1]{%
  \begingroup
  \renewcommand\thefootnote{}\footnote{#1}%
  \addtocounter{footnote}{-1}%
  \endgroup
}

\input{defs}

\begin{document}

\title{Trick-GS: A Balanced Bag of Tricks for Efficient Gaussian Splatting}

\author[]{Anil Armagan}
\author[]{Albert Sa\`a-Garriga}
\author[]{Bruno Manganelli}
\author[]{Mateusz Nowak}
\author[]{M. Kerim Yucel}
\vspace{-0.5em}
\affil[]{Samsung R\&D Institute UK (SRUK)}

\maketitle

\begin{abstract}

Gaussian splatting (GS) for 3D reconstruction has become quite popular due to their fast training, inference speeds and high quality reconstruction. However, GS-based reconstructions generally consist of millions of Gaussians, which makes them hard to use on computationally constrained devices such as smartphones. In this paper, we first propose a principled analysis of advances in efficient GS methods. Then, we propose Trick-GS, which is a careful combination of several strategies including (1) progressive training with resolution, noise and Gaussian scales, (2) learning to prune and mask primitives and SH bands by their significance, and (3) accelerated GS training framework. Trick-GS takes a large step towards resource-constrained GS, where faster run-time, smaller and faster-convergence of models is of paramount concern. Our results on three datasets show that Trick-GS achieves up to 2$\times$ faster  training, 40$\times$ smaller  disk  size and  2$\times$ faster  rendering  speed compared to vanilla GS, while having comparable accuracy.
\end{abstract}

3D reconstruction, Gaussian splatting, 3DGS.

\input{introduction}
\input{related_work_short}

\blfootnote{
Copyright 2025 IEEE. Published in ICASSP 2025 – 2025 IEEE International Conference on Acoustics, Speech and Signal Processing (ICASSP), scheduled for 6-11 April 2025 in Hyderabad, India. Personal use of this material is permitted. However, permission to reprint/republish this material for advertising or promotional purposes or for creating new collective works for resale or redistribution to servers or lists, or to reuse any copyrighted component of this work in other works, must be obtained from the IEEE. Contact: Manager, Copyrights and Permissions / IEEE Service Center / 445 Hoes Lane / P.O. Box 1331 / Piscataway, NJ 08855-1331, USA. Telephone: + Intl. 908-562-3966.}
\input{methodology}

\input{experiments}

\input{conclusion}

\clearpage

\bibliographystyle{IEEEbib}
\bibliography{strings,main}

\end{document}

%% file: defs.tex
\usepackage[normalem]{ulem}

\newcommand{\figwidth}{0.2}

\newcolumntype{F}[1]{>{\raggedright\arraybackslash\hspace{0pt}}p{#1}}%
\newcolumntype{T}[1]{>{\centering\arraybackslash\hspace{0pt}}p{#1}}%
\newcolumntype{P}[1]{>{\centering\arraybackslash\hspace{0pt}\vspace{0pt}}p{#1}}

\def\BibTeX{{\rm B\kern-.05em{\sc i\kern-.025em b}\kern-.08em
    T\kern-.1667em\lower.7ex\hbox{E}\kern-.125emX}}

\newcommand{\firstcell}[1]{{\cellcolor{blue!30} #1}}
\newcommand{\secondcell}[1]{{\cellcolor{blue!20} #1}}
\newcommand{\thirdcell}[1]{{\cellcolor{blue!10} #1}}

\DeclareRobustCommand\onedot{\futurelet\@. }
\def\eg{{e.g}\onedot}

\def\wrt{w.r.t\onedot}

\newcommand{\cmark}{\ding{51}}%
\newcommand{\xmark}{\ding{55}}%

\usepackage{mathtools}
\usepackage{xspace}

%% file: introduction.tex
\section{Introduction} \label{introduction}
\noindent 3D reconstruction is a long standing problem in computer vision, with applications in robotics \cite{adamkiewicz2022vision}, VR \cite{li2023instant} and multimedia \cite{skartados2024finding}. It is a notorious problem since it requires lifting 2D images to 3D space in a dense, accurate manner. Gaussian splatting (GS) \cite{kerbl3Dgaussians} is the current state-of-the-art approach for 3D reconstruction, owing to its fast rendering, convergence and high accuracy characteristics. However, it tends to have a large disk size due to the usage of millions of Gaussians, take longer to train compared to hash-grid based NeRF methods \cite{muller2022instant} and can still benefit from improvements across various axes.

In this paper, with the aim of taking a step towards device friendly GS, we perform a principled analysis of various approaches that produce efficient \cite{lee2024compact, fang2024minisplatting} GS pipelines. These include Gaussian pruning \cite{lee2024compact}, progressive training \cite{girish2024eagles, yang2024spec}, blurring  \cite{girish2024eagles} and spherical harmonic masking \cite{wang2024end}. After a careful exploration, we then propose \textbf{\textit{Trick-GS}}, a GS method that is created by systematically combining these approaches. \textit{Trick-GS} has comparable accuracy to vanilla GS and other compact GS methods on multiple benchmark datasets, and enjoys up to 2 $\times$ faster training, 40 $\times$ smaller disk size and 2 $\times$ faster rendering speed compared to other methods. Furthermore, \textit{Trick-GS} is flexible, where one can tune the design to prioritize different aspects of the pipeline, such as convergence and rendering speed, accuracy and disk size. 

%% file: related_work_short.tex
\section{Related Work} \label{related_work}
\noindent \textbf{Storage efficient GS.} Gaussian primitives in GS are represented with 59 attributes. Storing all Gaussians, therefore, might lead to large disk size consumption. There are various methods achieving storage efficient GS, via reducing the number of Gaussians by pruning subject to various criteria \cite{fan2024lightGaussian,hanson2024pup3dgs,morgenstern2024compact,papantonakis2024reducing,ali2024trimmingfat} and reducing the number of attributes in each Gaussian by masking/compressing spherical harmonic bands or color parameters \cite{papantonakis2024reducing, lee2024compact, fan2024lightGaussian}. Applying Vector Quantization \cite{lee2024compact, fan2024lightGaussian, niedermayr2024compressed} and bit-quantization \cite{wang2024end, niedermayr2024compressed} have also found use. Using anchors to derive the Gaussians \cite{lu2024scaffold, chen2024hac, wang2024contextgs, liu2024compgs}, using self-organizing mapping between 2D grids and Gaussian attributes \cite{morgenstern2024compact} or adopting representations such as Octree \cite{ren2024octree} or tri-planes \cite{wu2024implicit} are other example methods for storage efficient GS.

\noindent \textbf{Fast-converging GS.} Most GS methods need fast convergence as they require per-scene training. \cite{mallick2024taming3dgs} speeds-up training speed by separating 0th and higher SH bands into different tensors and load into rasterizer separately. \cite{durvasula2023distwar} introduce ways to accelerate atomic operations prevalent in gradient computations.

\noindent \textbf{Fast-rendering GS.} Despite the already fast rendering speed of GS, new GS methods have managed to improve the rendering speed further. \cite{jo2024unnecessary} removes Gaussians via offline clustering. \cite{lee2024GSCore} proposes hardware accelaration units to focus on Gaussian shape-aware intersection tests, hierarchical sorting, and sub-tile skipping.

%% file: methodology.tex
\input{figs/qualitative.tex}

\section{Methodology}
\label{methodology}
\subsection{Preliminaries}
3DGS reconstructs a scene by fitting a collection of 3D Gaussian primitives, which can be efficiently rendered in a differentiable volume splatting rasterizer by extending EWA volume splatting \cite{zwicker2001ewa}. A scene represented with 3DGS typically consists of hundreds of thousands to millions of Gaussians, where each 3D Gaussian primitive consists of 59 parameters. The 2D Gaussian $G_{i}'$ projected from a 3D Gaussian $G_i$ is defined as
\begin{equation}
\label{eq:2d-Gaussian-proj}
   G_{i}'(x) = e^{-\frac{1}{2}(x-\mu_{i}')^T{\Sigma_{i}'}^{-1}(x-\mu_{i}')},\vspace{-0.6em}
\end{equation}
\noindent
where \(x\) denotes the position vector, \(\mu_{i}\) represents 3D position, and \(\Sigma_{i}\) is the 3D covariance matrix which is later parameterized using scaling matrix \(S\) and a rotation matrix that is represented in quaternions \(q\). The symbol $'$ represents 2D reprojection of corresponding parameters. Each Gaussian primitive has an opacity (\(\alpha \in [0, 1]\)), a diffused color and a set of spherical harmonics (SH) coefficients, typically consisting of 3-bands, to represent view-dependent colors. Color \(C\) of each pixel in the image plane is later determined by \(N\) Gaussians contributing to that pixel with blending the colors in a sorted order.
\begin{equation}
\label{colorblending}
    C = \sum_{i \in N} c_i \alpha_i  \prod_{j=1}^{i-1} (1 - \alpha_j),\vspace{-0.6em}
\end{equation}
\noindent
where \(c_i\) and \(\alpha_i\) are the view-dependent color and opacity of a Gaussian. 3DGS is trained using a reconstruction loss with L1 and SSIM losses, and Gaussians are pruned and densified based on their opacities and gradient magnitudes during training. We refer to original paper for more details.

\subsection{Tricks for Learning Efficient 3DGS Representations}
In this section we give details of strategies we adopt to obtain a balanced 3DGS representation in terms of disk storage space, training time and rendering time. 
\subsubsection{\textbf{Pruning with Volume Masking}}
\label{sec:volume-masking}
Gaussians with low scale tend to have minimal impact on the overall quality, therefore we adopt a strategy \cite{morgenstern2024compact} that learns to mask and remove such Gaussians. In short, $N$ binary masks, $M\in \{0,1\}^{N}$, are learned for $N$ Gaussians and applied on its opacity, $\alpha\in [0,1]^{N}$, and non-negative scale attributes, $s\in \mathbb{R}_{+}^{N\times3}$ by introducing a mask parameter, $m\in \mathbb{R}^{N}$, for the Gaussian scene representation. The learned masks are then thresholded to generate hard masks $M$.
\begin{equation}
\small
    M_n = \operatorname*{sg}(\mathds{1}(\sigma(m_n) > \epsilon) - \sigma(m_n)) + \sigma(m_n),\vspace{-0.4em}
\end{equation}
\noindent
where $\epsilon$ is the masking threshold, $\operatorname*{sg}$ is the stop gradient operator, and $\mathds{1}(\cdot)$ and $\sigma(\cdot)$ are indicator and sigmoid function, respectively for the $n^{th}$ Gaussian. In training, these Gaussians are pruned at densification stage using these masks and are also pruned at every $k_{m}$ iteration after densification stage.

\subsubsection{\textbf{Pruning with Significance of Gaussians}} In parallel to scale and opacity based pruning, another axis is to find and remove Gaussians with little overall impact. For further pruning, we adopt a strategy~\cite{fan2024lightGaussian} where the impact of a Gaussian is determined by considering how often it is hit by a ray. More concretely, the so called significance score is calculated as $\mathbbm{1}(\text{G}(X_j), r_i)$ where $\mathbbm{1}(\cdot)$ is the indicator function, $r_i$ is ray $i$ for every ray in the training set and $X_{j}$ is the times Gaussian $j$ hit $r_{i}$. This score is then harmonized with the Gaussian's volume and opacity as in Eq.~\ref{eq:significance-score}.

\begin{equation}
\label{eq:significance-score}
\small
\text{GS}_{\text{j}} = \sum_{i=1}^{MHW} \mathbbm{1}(\text{G}(X_j), r_i) \cdot \alpha_j \cdot {\large{\gamma}}(\boldsymbol{\Sigma}_j),\vspace{-0.4em}
\end{equation}
\noindent
where $j$ is the Gaussian index, $i$ is the pixel,   \raisebox{2pt}{$\large{\gamma}$}$(\boldsymbol{\Sigma}_j)$ is the Gaussian's volume, $M$, $H$, and $W$ represents the number of training views, image height, and width, respectively. This volume calculation is then refined by normalizing the volume of a Gaussian by the volume of 90\% largest of all sorted Gaussians to avoid excessive floating Gaussians. This significance score is used to prune a preset percentile of Gaussians during training. Since it is a costly operation, we apply this pruning $k_{sg}$ times during training with a decay factor considering the percentile removed in the previous round. 

\input{tables/mip_tandt_table.tex}

\subsubsection{\textbf{Spherical Harmonic (SH) Masking.}} SHs are used to represent view dependent color for a Gaussian, however, one can appreciate that not all Gaussians will have the same levels of varying colors depending on the scene, which provides a further pruning opportunity. We adopt a strategy \cite{wang2024end} where SH bands are pruned based on a mask learned during training, and unnecessary bands are removed after the training is complete. Specifically, each Gaussian learns a mask per SH band. SH masks are calculated as in Eq. \ref{eq:sh-masks} and SH values for the $i^{th}$ Gaussian for the corresponding SH band, $l$, is set to zero if its hard mask, $M_{{sh}_{i}}^{l}$, value is zero and unchanged otherwise.
\begin{equation}
\label{eq:sh-masks}
\small
    M_{{sh}_{i}}^{l} = \texttt{sg}\left(\mathbbm{1}\left(\sigma(m_{{sh}_{i}}^{l}) > \epsilon_{sh}\right) - \sigma(m_{{sh}_{i}}^{l})\right) + \sigma(m_{{sh}_{i}}^{l}),\vspace{-0.4em}
\end{equation}
\noindent
where $m_{i}^{l} \in \left(0,1\right)$, $M_{{sh}_{i}}^{l} \in \{0,1\}$. Finally, each masked  view-dependent color is defined as $\mathbf{\hat{c}}_{i}^{l} = M_{{sh}_{i}}^{l} \mathbf{c}_i^{l}$ where $\mathbf{c}_i^{l}\in\mathbb{R}^{(2l+1)\times3}$. Masking loss for each degree of SH is weighted by its number of coefficients, since the number of SH coefficients vary per SH band.

\subsubsection{\textbf{Progressive Training.}}
Progressive training of Gaussians refers to starting from a coarser, less detailed image representation and gradually changing the representation back to the original image. This approach has been proven to work as a regularization scheme~\cite{yang2024spec, girish2024eagles} for densification and pruning of Gaussians as the initialization of the Gaussians from an \(SfM\) point-cloud can be sub-optimal. We use three different strategies in this aspect.

\paragraph{\textbf{Progressive training by blurring.}}
\label{sec:progressive-blur} Gaussian blurring is used to change the level of details in an image~\cite{girish2024eagles}. Kernel size is progressively lowered at every $k_{b}$ iteration based on a decay factor. This strategy helps to remove floating artifacts from the sub-optimal initialization of Gaussians and serves as a regularization to converge to a better local minima. It also significantly impacts the training time since a coarser scene representation requires less number of Gaussians to represent the scene.

\paragraph{\textbf{Progressive training by resolution.}}
Another strategy to mitigate the over-fitting on training data is to start with smaller images and progressively increase the image resolution during training to help learning a broader global information~\cite{yang2024spec, girish2024eagles}. This approach specifically helps to learn finer grained details for pixels behind the foreground objects. Resolution of an image during training calculated as in Eq.\ref{eq:progressive-res}.\vspace{-0.4em}
\begin{equation}
\small
\label{eq:progressive-res}
    res(i) = \min(\lfloor res_s + (res_e - res_s) \cdot i/{\tau_{res}} \rceil, res_e),\vspace{-0.4em}
\end{equation}
\noindent 
where $res(i)$ is the image resolution at the $i^{th}$ training iteration, $res_s$ is the starting image resolution, $res_e$ is the full training image resolution, and $\tau_{res}$ is the threshold iteration.

\paragraph{\textbf{Progressive training by scales of Gaussians.}}
Another strategy is to focus on low-frequency details during the early stages of the training by controlling the low-pass filter in the rasterization stage. Some Gaussians might become smaller than a single pixel if each 3D Gaussian is projected as Eq.~\ref{eq:2d-Gaussian-proj}, which results in artifacts. Therefore, the covariance $\Sigma_{i}'$ is replaced with $\Sigma_{i}'+sI$ by adding a small value to the diagonal element~\cite{kerbl3Dgaussians} where $s=0.3$ and $I$ is an identity matrix.

~\cite{jung2024relaxing} progressively changes $s$ for each Gaussian during the optimization to ensure the minimum area that each Gaussian covers in the screen space. Using a larger $s$ in the beginning of the optimization enables Gaussians receive gradients from a wider area and therefore the coarse structure of the scene is learned efficiently. The value $s$ ensures the projected Gaussians area to be larger than $9{\pi}s$, please refer to~\cite{jung2024relaxing} for a proof, and $s$ is defined as $s = HW/9\pi N$, where $N$, $H$ and $W$ indicate the number of Gaussians, image height and width, respectively. This strategy is lower bounded with a default value of $s=0.3$ as the number of Gaussians increases. We start the optimization with the 20\% of the 3D points from the \(SfM\) initialization that has the lowest error to help for a more efficient training. 

\subsubsection{\textbf{Accelerated training for 3DGS}}
We adopt a strategy \cite{mallick2024taming3dgs} that is more focused on the training time efficiency. We follow \cite{mallick2024taming3dgs} on separating higher SH bands from the $0^{th}$ band within the rasterization, thus lower the number of updates for the higher SH bands than the diffused colors. SH bands (45 dims) cover a large proportion of the these updates, where they are only used to represent view-dependent color variations.~\cite{mallick2024taming3dgs} modifies the rasterizer to split SH bands from the diffused color. SH bands are updated every $16$ iterations where as diffused color is updated at every step. To make the training faster we further modify the $SSIM$ loss calculation~\cite{mallick2024taming3dgs} with optimized CUDA kernels. $SSIM$ is configured to use $11\times11$ Gaussian kernel convolutions by standard, where as an optimized version is obtained by replacing the larger 2D kernel with two smaller 1D Gaussian kernels. SSIM metric is calculated with a fused kernel from the output of convolutions. Applying less number of updates for higher SH bands and optimizing the $SSIM$ loss calculation have a negligible impact on the accuracy, while it significantly helps to speed-up the training time as shown by our experiments. 

Finally, we can define our loss as
\begin{equation}
\small
\label{loss3dgs}
\mathcal{L}=(1-\lambda_{SSIM}) \mathcal{L}_1+ \lambda_{SSIM} \mathcal{L}_{\text{D-SSIM}} + \lambda_{m} \mathcal{L}_{\text{m}} + \lambda_{sh} \mathcal{L}_{\text{sh}} 
\end{equation}
\noindent
since only Gaussian and SH masking are learnable. Please see the corresponding papers~\cite{morgenstern2024compact,fan2024lightGaussian} for the definitions of $L_m$ and $L_{sh}$.

%% file: figs/qualitative.tex
\begin{figure*}[t]
\centering
\captionsetup[subfigure]{labelformat=empty}
\vspace{-0.4em}
\subfloat[]{\includegraphics[width=\figwidth\linewidth]{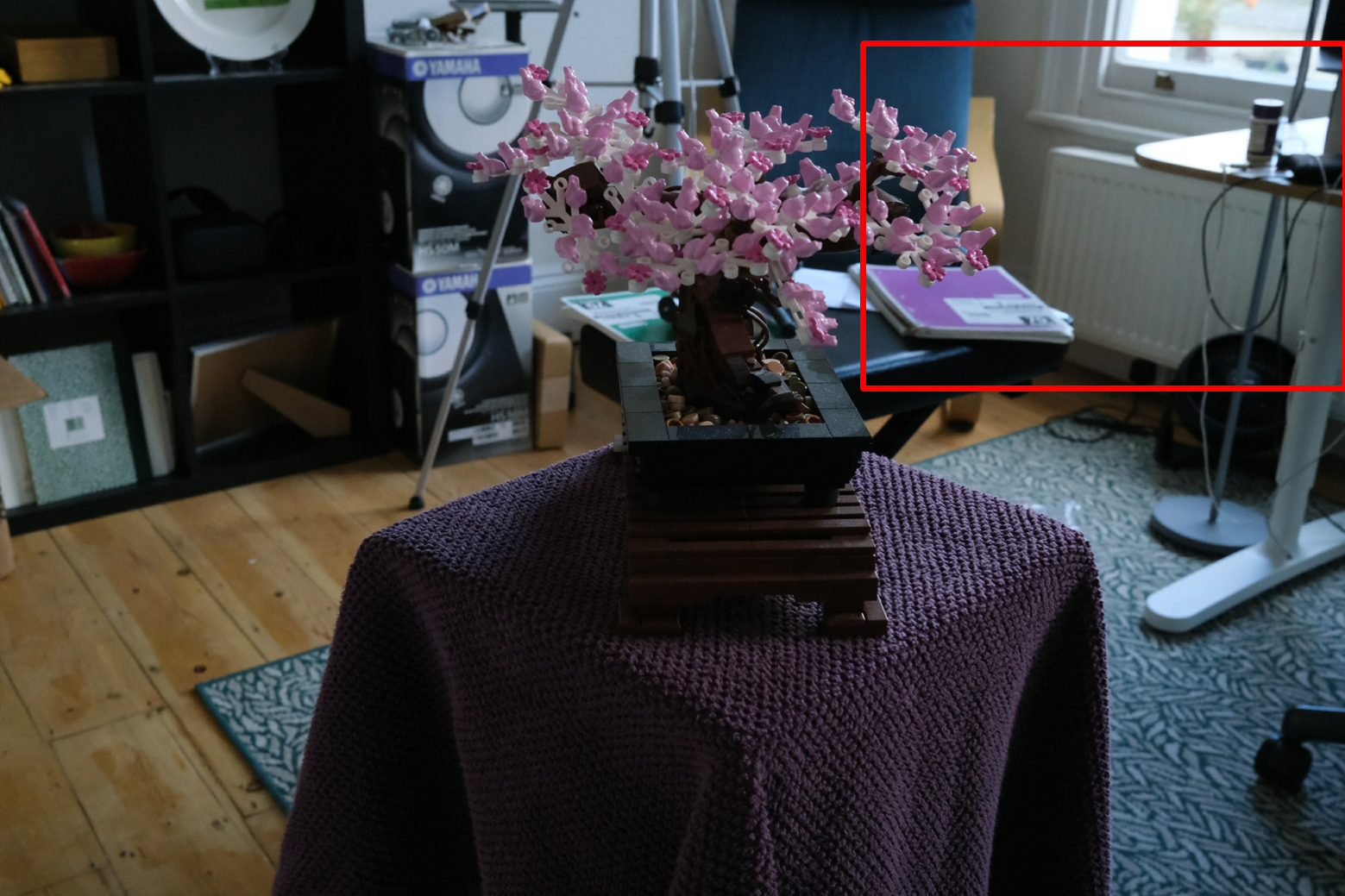}} 
\subfloat[]{\includegraphics[width=\figwidth\linewidth]{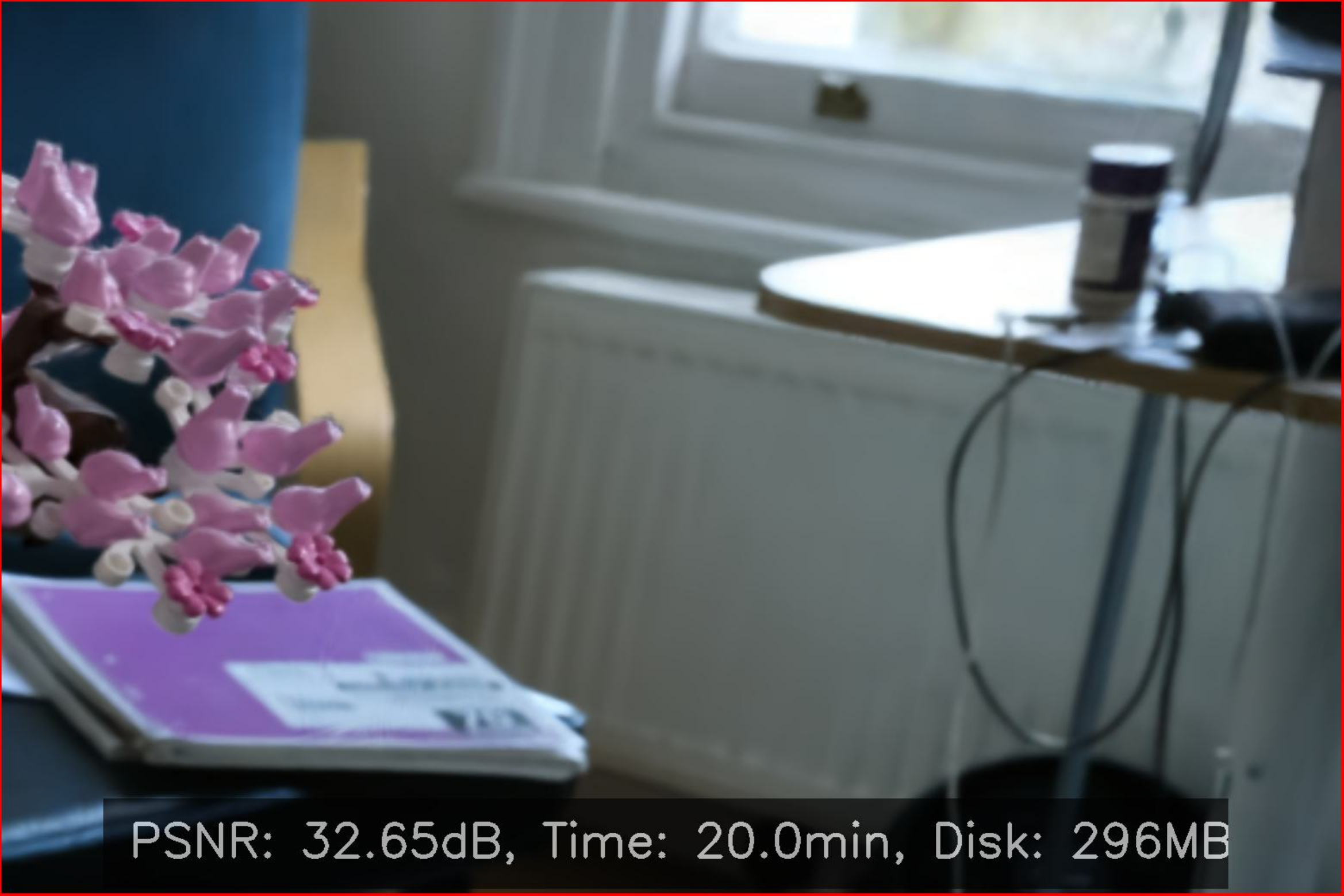}} 
\subfloat[]{\includegraphics[width=\figwidth\linewidth]{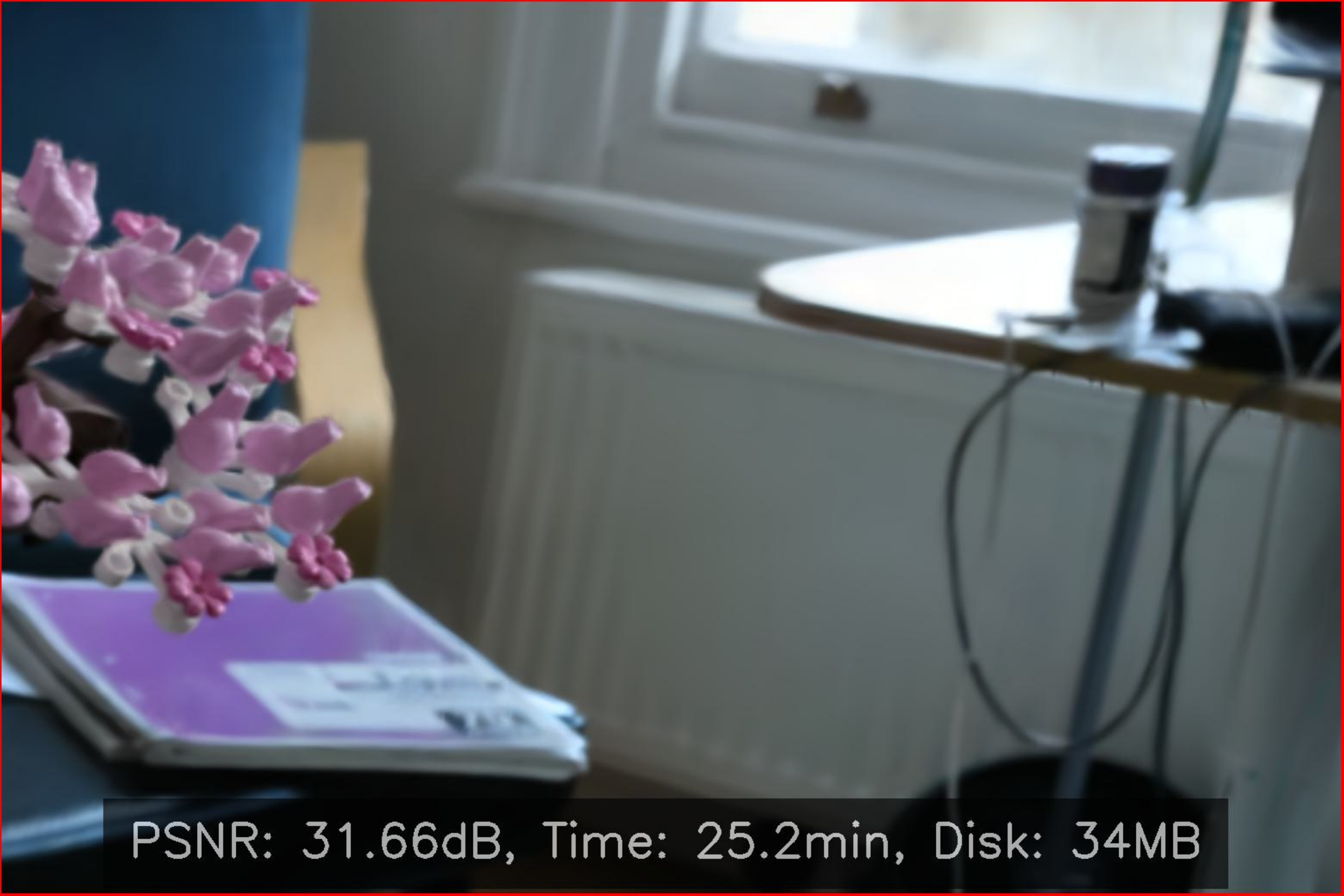}} 
\subfloat[]{\includegraphics[width=\figwidth\linewidth]{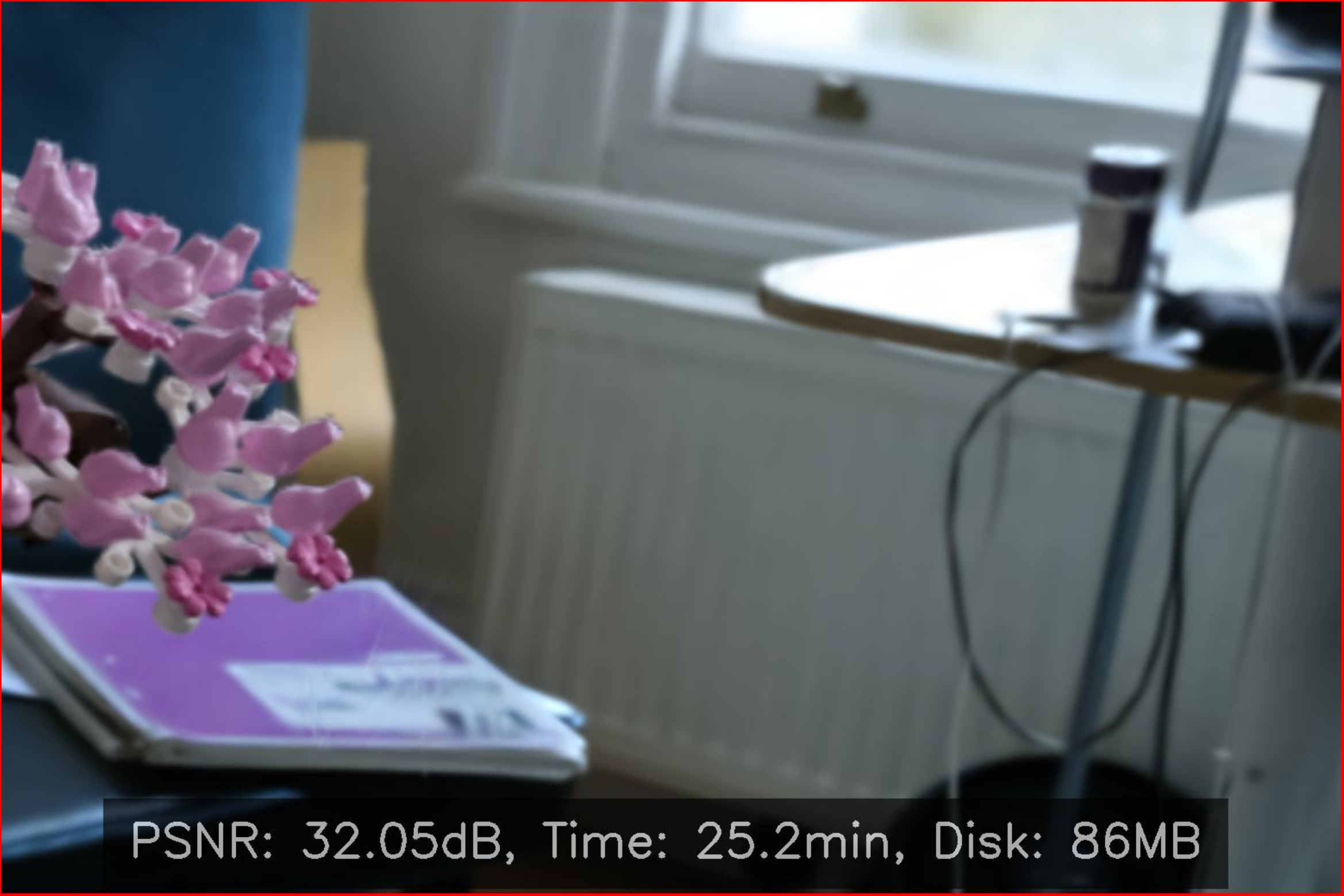}}
\subfloat[]{\includegraphics[width=\figwidth\linewidth]{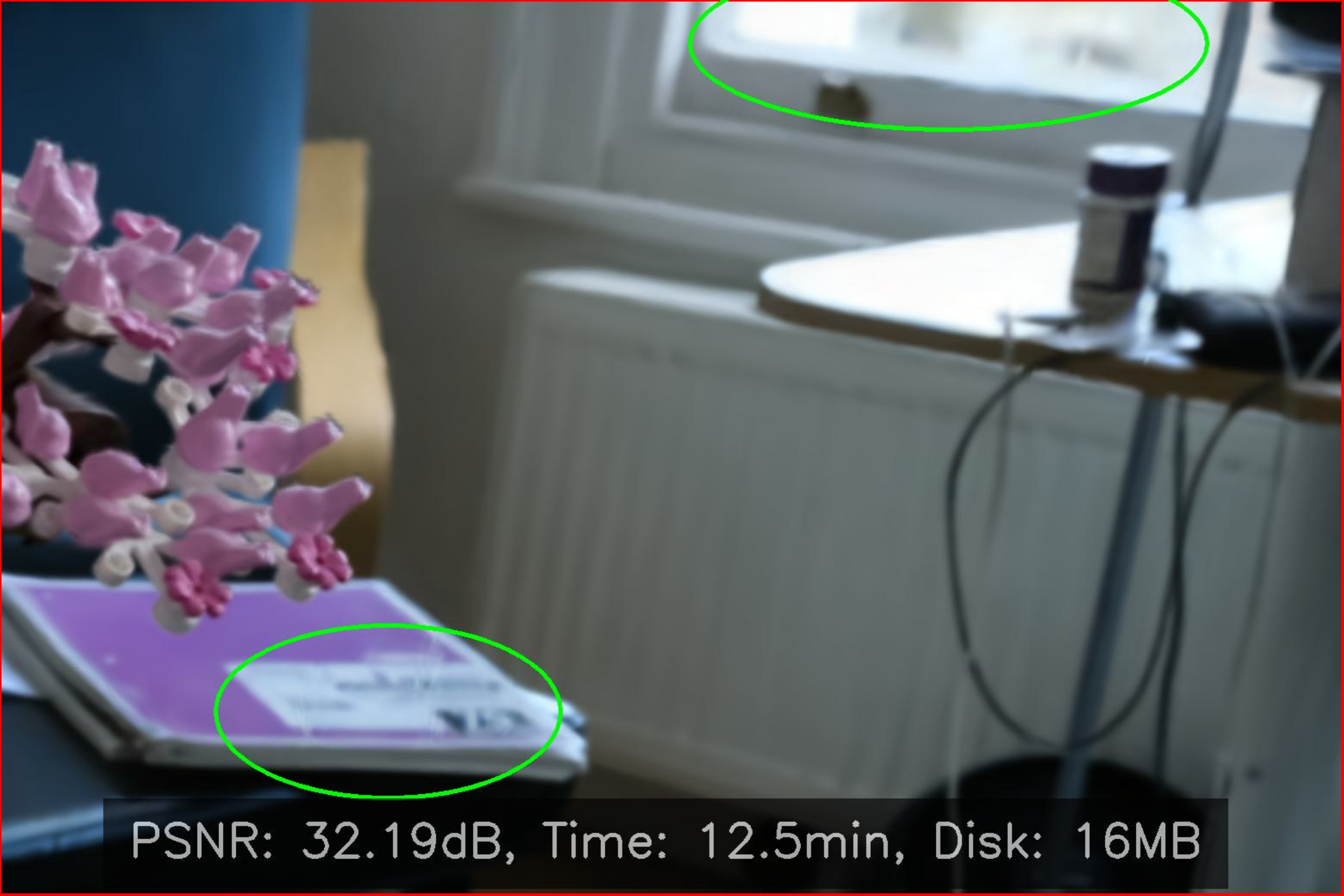}} \\
\vspace{-2.3em}
\subfloat[(a) GT]{\includegraphics[width=\figwidth\linewidth]{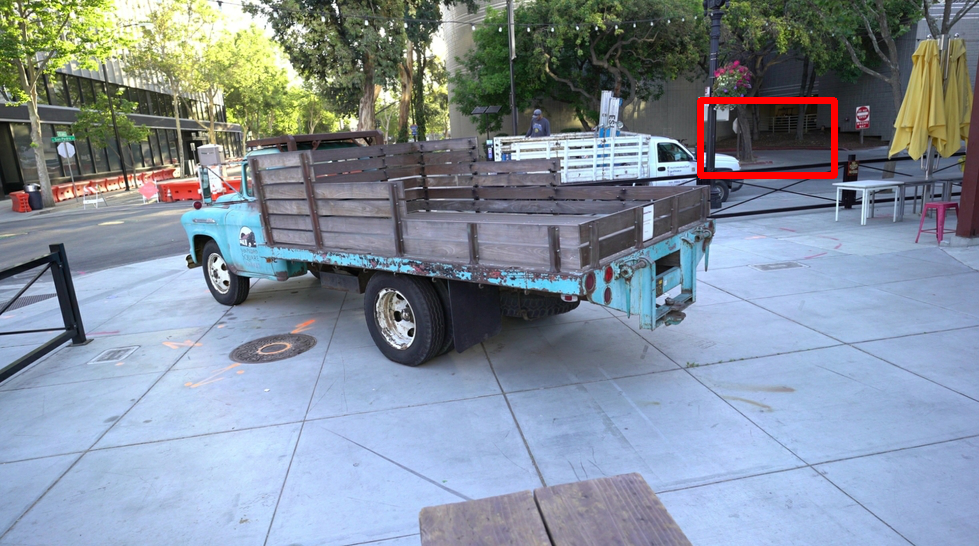}} 
\subfloat[(b) 3DGS~\cite{kerbl3Dgaussians}]{\includegraphics[width=\figwidth\linewidth]{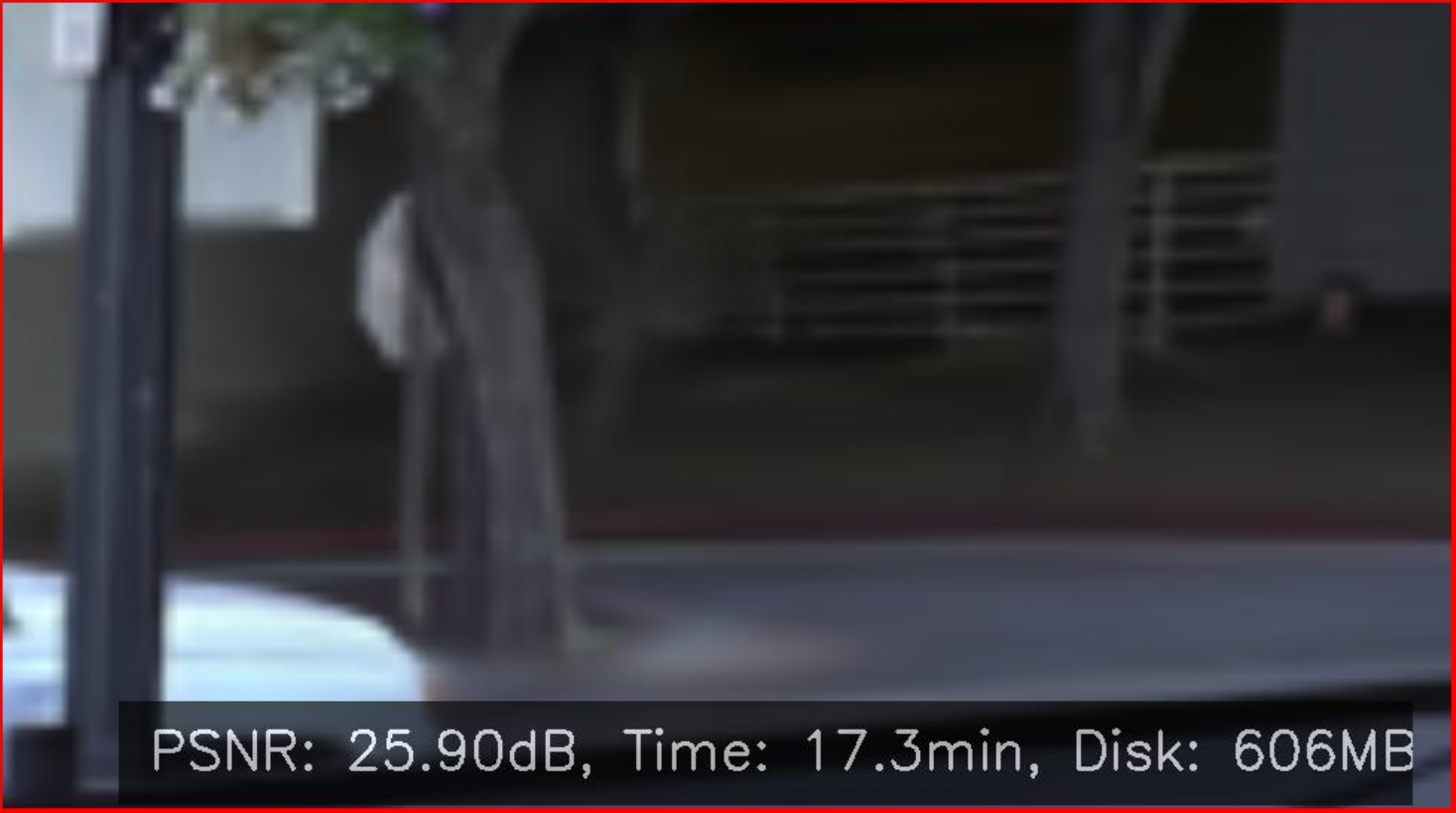}} 
\subfloat[(c) Compact-GS~\cite{lee2024compact}]{\includegraphics[width=\figwidth\linewidth]{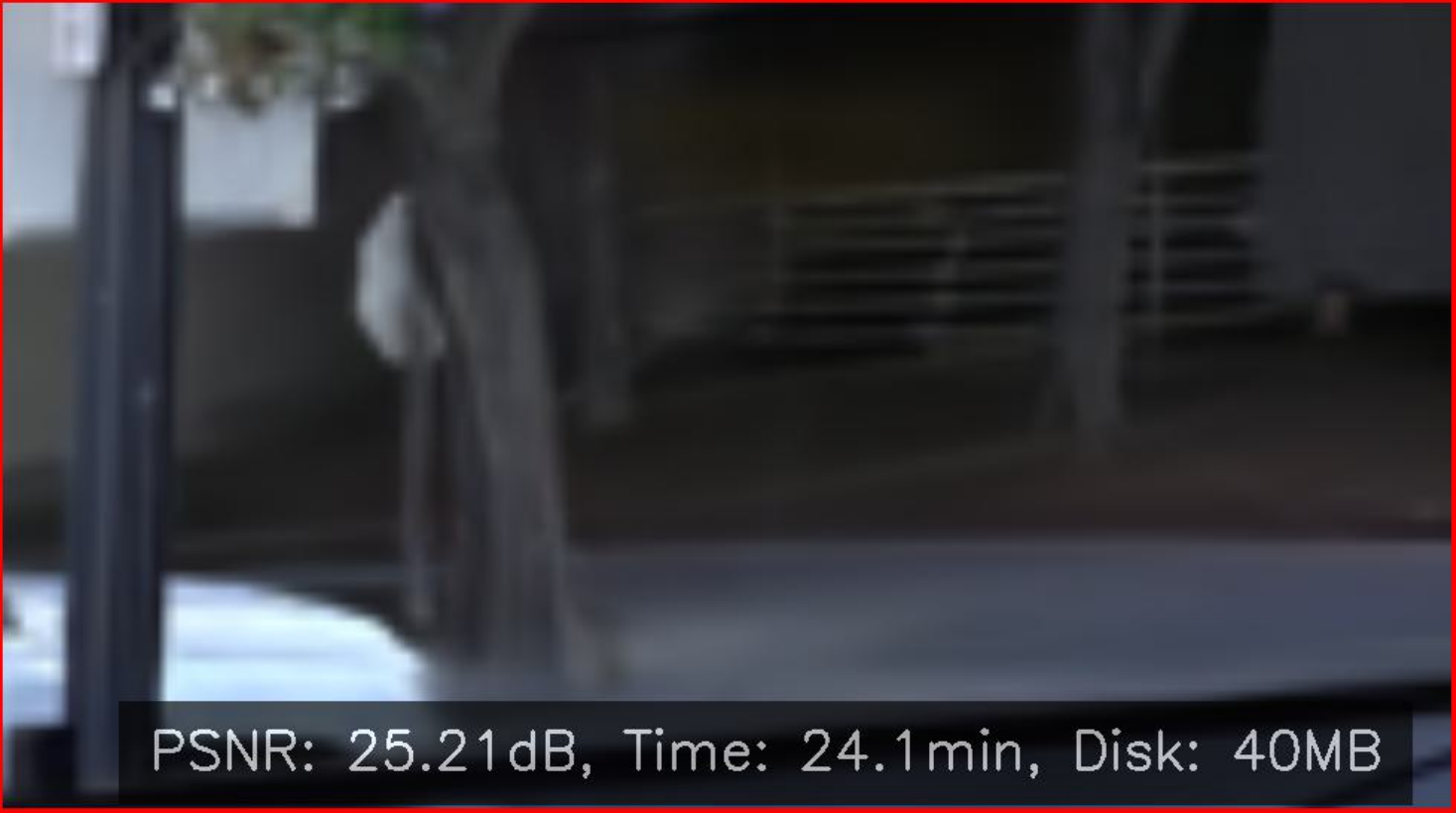}} 
\subfloat[(d) Mini-Splatting~\cite{fang2024minisplatting}]{\includegraphics[width=\figwidth\linewidth]{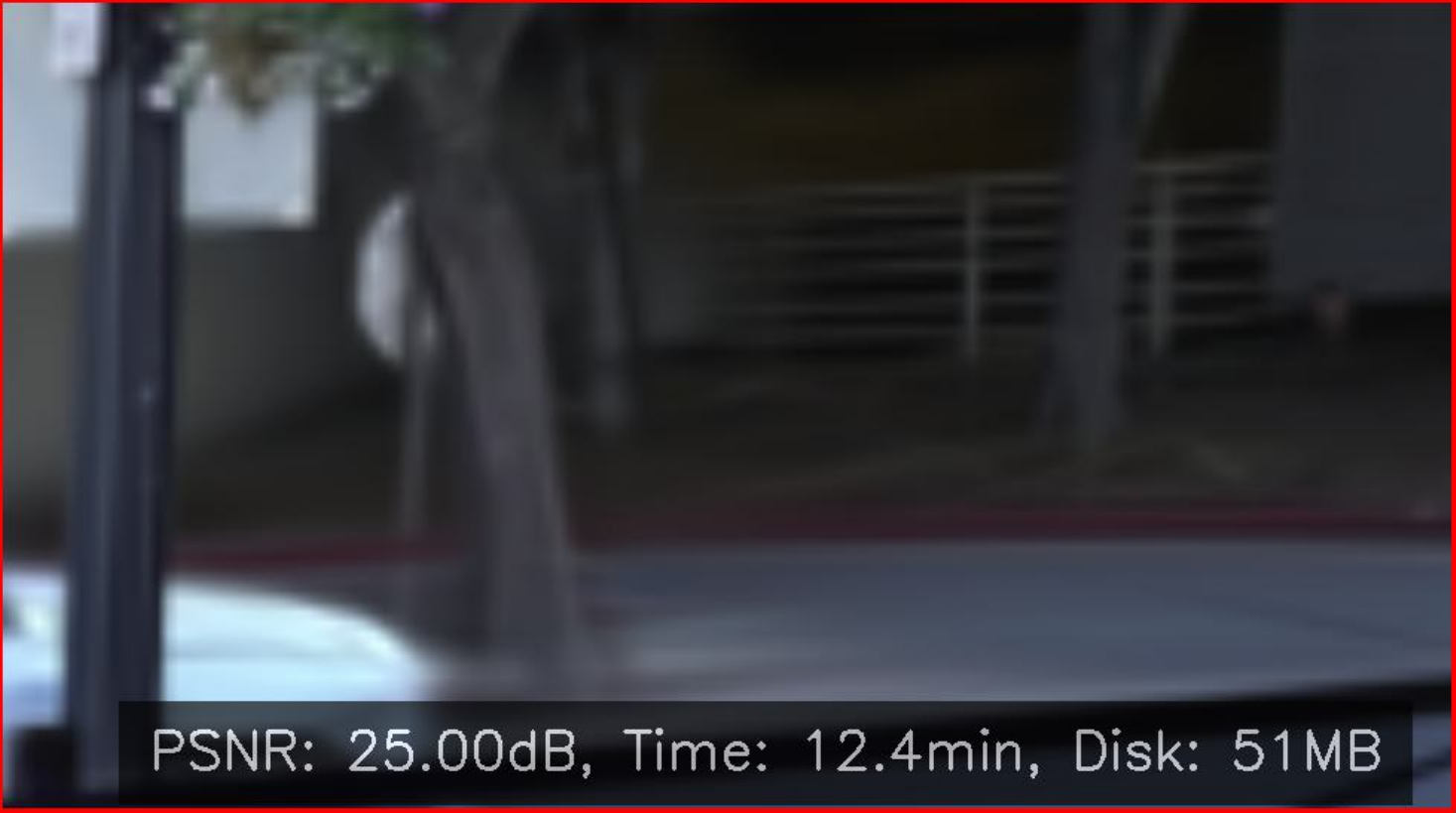}}
\subfloat[(e) Trick-GS (Ours)]{\includegraphics[width=\figwidth\linewidth]{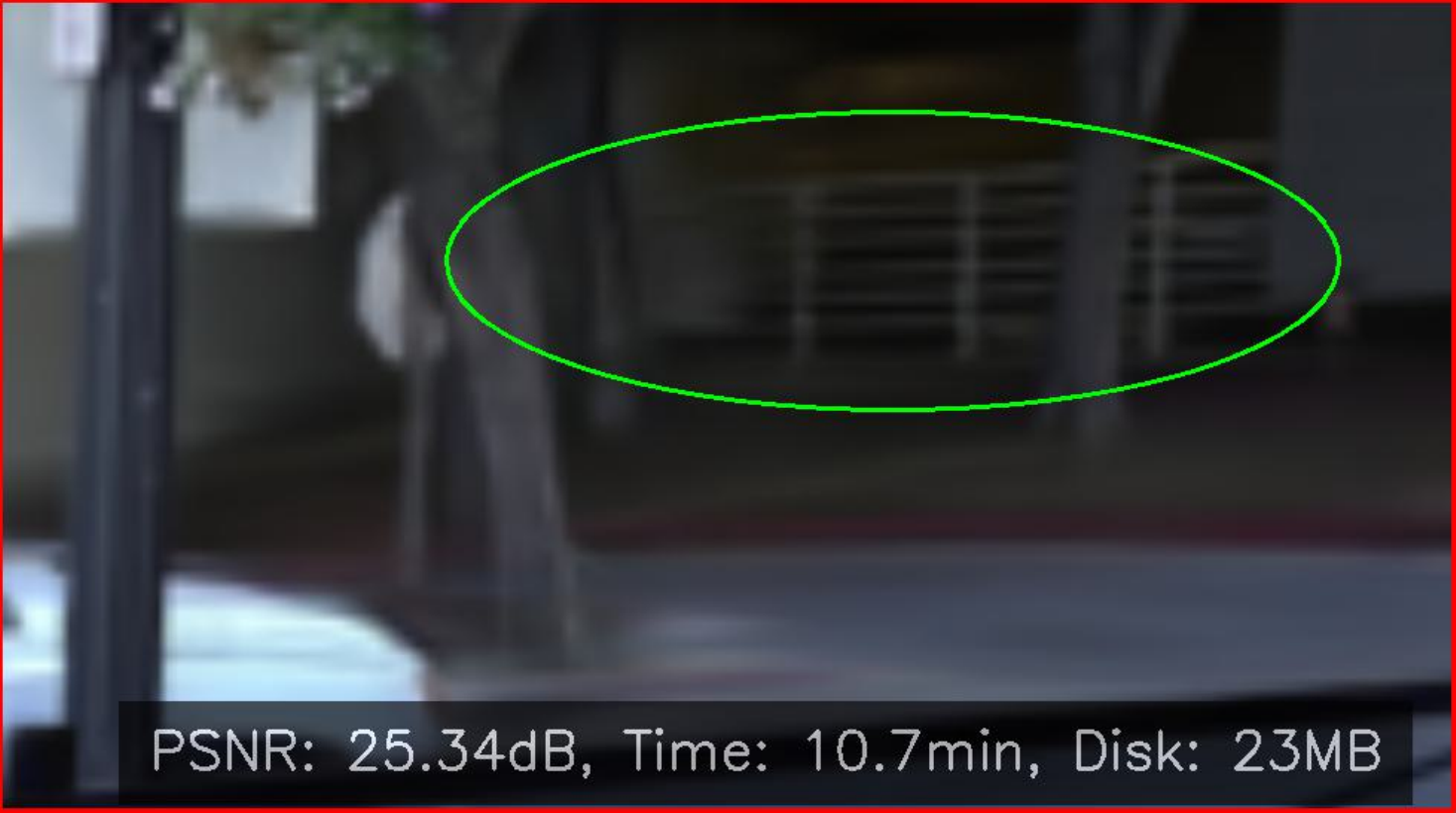}} 

\caption{Qualitative comparison of the methods. Our method can recover more consistent text and background (top), or better structure metallic fence (bottom) while keeping low training time and great compression rates. We show zoomed prediction images (b-e) for the cropped areas with red rectangles as in ground-truth image (a) and green circles for improvements. }
\label{fig:visual_comparison}
\end{figure*} 

%% file: tables/mip_tandt_table.tex
\begin{table*}[ht!]
\centering
\resizebox{1.0\linewidth}{!}{
\begin{tabular}{l|cccT{0.07\textwidth}T{0.05\textwidth}cT{0.08\textwidth}T{0.06\textwidth}|cccT{0.07\textwidth}T{0.05\textwidth}cT{0.08\textwidth}T{0.06\textwidth}}
\toprule
\rowcolor{lightgray} Dataset      & \multicolumn{8}{c}{Mip-NeRF 360}                & \multicolumn{8}{c}{Tanks\&Temples}              \\
\hline
Method & PSNR ↑ & SSIM ↑ & LPIPS ↓ & Storage ↓ (MB) & Time ↓ (min) &
FPS ↑ & Max-\#GS ↓ $\times1000$ & \#GS ↓ $\times1000$ & PSNR ↑ & SSIM ↑ & LPIPS ↓ & Storage ↓ (MB) & Time↓ (min) &
FPS ↑ & Max-\#GS ↓ $\times1000$ & \#GS ↓
$\times1000$ \\ 
\hline
\rowcolor{lightgray} \textsuperscript{*}3DGS~\cite{kerbl3Dgaussians} & 27.21 & 0.815 & 0.214 & 734 & 41.52 & 134 & - & - & 23.14 & 0.841 & 0.183 & 411 & 26.90 & 154 & - & -\tabularnewline
\textsuperscript{*}INGP-base \cite{muller2022instant} & 25.30	& 0.671	& 0.371	& 13 & 5.62	& 12 & - & - & 21.72 & 0.723 & 0.330 & 13 & 5.43 & 17 & - & -\tabularnewline
\rowcolor{lightgray} \textsuperscript{*}INGP-big \cite{muller2022instant} & 25.59	& 0.699	& 0.331	& 48 & 7.50	& 9 & - & - & 21.92 & 0.745 & 0.305 & 48 & 6.98 & 14 & - & -\tabularnewline
\textsuperscript{*}Mini-Splatting~\cite{fang2024minisplatting} & 27.34 & 0.822 & 0.217 & - & - & - & - & 490 & 23.18 & 0.835 & 0.202 & - & - & - & - & 200\tabularnewline
\rowcolor{lightgray} \textsuperscript{*}Compact-GS~\cite{lee2024compact} & 27.08 & 0.798 & 0.247 & 49	& 33.10	& 128 & - & - & 23.32 & 0.831 & 0.201 & 39.4 & 18.30 & 185 & - & -\tabularnewline

\hline\hline

3DGS~\cite{kerbl3Dgaussians} & \firstcell{27.56} & \secondcell{0.818} & \firstcell{0.202} & 770 & 23.83 & 142 & 3265 & 3255 & \firstcell{23.67} & \firstcell{0.845} & \firstcell{0.178} & 431 & 14.69 & 172 & 1824 & 1824\tabularnewline

\rowcolor{lightgray} Mini-Splatting~\cite{fang2024minisplatting} & \secondcell{27.25} & \firstcell{0.820} & \secondcell{0.219} & 118 & \thirdcell{21.92} & \firstcell{364} & 4235 & \secondcell{496} & 23.23 & \secondcell{0.833} & \secondcell{0.202} & 48 & \thirdcell{12.34} & \firstcell{452} & 4291 & \firstcell{203}\tabularnewline

Compact-GS~\cite{lee2024compact} & 26.95 & 0.797 & \thirdcell{0.245} & \thirdcell{47} & 41.74 & 128 & \thirdcell{2567} & 1406 & \thirdcell{23.29} & \thirdcell{0.830} & \secondcell{0.202} & \thirdcell{38} & 21.15 & 192 & \thirdcell{1467} & 834\tabularnewline

\hline\hline
\rowcolor{lightgray} Trick-GS-small & 26.714	& 0.780	& 0.285	& \firstcell{19}	& \firstcell{12.42}	& \secondcell{256} & \firstcell{1033}	& \firstcell{392} & 23.21 & 0.818	& 0.232	& \firstcell{11}	& \firstcell{8.79}	& \secondcell{342} & \firstcell{570} & \secondcell{220} \tabularnewline

Trick-GS & \thirdcell{27.16} & \secondcell{0.802} & \thirdcell{0.245} & \secondcell{39} & \secondcell{15.41} & \thirdcell{222}  & \secondcell{1369} & \thirdcell{830} & \secondcell{23.48} & \thirdcell{0.830} & \thirdcell{0.209} & \secondcell{20} & \secondcell{10.56} & \thirdcell{298} & \secondcell{696} & \thirdcell{443}
\\\bottomrule
\end{tabular}}
\caption{Quantitative evaluation on MipNeRF 360 and Tanks\&Temples datasets. Results with marked '$^*$' method names are taken from the corresponding papers. Results between the double horizontal lines are from retraining the models on our system. We color the results with $1^{st}$, $2^{nd}$ and $3^{rd}$ rankings in the order of solid to transparent colored cells for each column. Trick-GS can reconstruct scenes with much lower training time and disk space requirements while not sacrificing on the accuracy.}
\label{tab:qual1}
\end{table*}

%% file: experiments.tex
\section{Experimental Results}
\label{experiments}
\subsection{Experimental Setup}
We follow the setup of \cite{kerbl3Dgaussians} on real-world scenes. 15 scenes from bounded and unbounded indoor/outdoor scenarios; nine from Mip-NeRF 360 \cite{MipNeRF360}, two (\textit{truck} and \textit{train}) from Tanks\&Temples \cite{TanksAndTemples} and two (\textit{DrJohnson} and \textit{Playroom}) from Deep Blending \cite{DeepBlending} datasets are used. \(SfM\) points and camera poses are used as provided by the authors \cite{kerbl3Dgaussians} and every 8\textsuperscript{th} image in each dataset is used for testing.  

Similar to \cite{kerbl3Dgaussians}, models are trained for 30K iterations and PSNR, SSIM and LPIPS~\cite{zhang2018unreasonable} are used for evaluation. We report training time using \textit{torch.cuda.Event} function including the time spent for the densification step, and FPS run-time over 50 runs using \textit{torch.utils.benchmark}. 

Note that \textit{Trick-GS} does not change the structure of 3DGS for efficiency, such as anchor points or explicit representations. Instead our method benefit from existing techniques mostly applied as a training strategy. We apply a simple post-processing by storing all Gaussian parameters with 16-bit precision using half-tensors after the training as we found the accuracy drop is insignificant. To make a fair comparison, we evaluate all methods on the same system using an \textit{NVIDIA RTX 3090}.

\input{tables/dp_table}

\subsection{Implementation Details}
We downsample images by 8$\times$ and gradually increase it with a logarithmic decay back to the original resolution until the iteration 19500. Similarly, we apply Gaussian blurring to the training images starting with a square kernel of $9\times9$ and $\sigma=2.4$ and gradually reduce it for every $k_{b}=100$ iterations until the iteration 19500. 

ABE split and progressive scale strategies~\cite{jung2024relaxing} are used for $10K$ iterations. Later standard densification \cite{kerbl3Dgaussians} is enabled at every $100$ iteration until iteration 15K.

Some of our strategies are pruning Gaussians at lower image scales even after the densification stops. Therefore, we enable the densification back between iterations $20K$ to $20.5K$ for every $100$ iterations to help the model to recover the loss that is from pruning of false-positive Gaussians. Learning rates $0.5$ and $0.05$ are used for learning the Gaussian and SH masks. The loss weights and the mask thresholds $\lambda_{m}=\lambda_{sh}=\epsilon_{m}=0.05$ and $\epsilon_{sh}=0.1$ are chosen. Gaussian masks are used to prune Gaussians at every densification step and for every $500$ iteration after the densification stops. 

Significance score based Gaussian pruning is applied $6$ times until iteration $22K$ with the first pruning rate set to 60\%, where a pruning decay factor of $0.7$ is used for the next iterations to reduce the number of pruned gaussians gradually in the later iterations. 

Different from the literature, we lower the blurring kernel size before we densify Gaussians. This helps to introduce an extra signal on the gradient of Gaussians that are more erroneous when the noise level is reduced. When downsampling and bluring is activated together, the kernel size is lowered by the scale of downsampling at the current stage, since the initial hyperparameter choice is based on original image resolutions.

\input{tables/ablation.tex}

\subsection{Performance Evaluation}

\input{figs/num_gaussians}

 Our most efficient model \textit{Trick-GS-small} improves over the vanilla 3DGS~\cite{kerbl3Dgaussians} by compressing the model size drastically, $23\times$, improving the training time and FPS by $1.7\times$ and $2\times$, respectively on three datasets. However, this results in slight loss of accuracy, and therefore we use late densification and progressive scale-based training with our most accurate model \textit{Trick-GS}, which is still more efficient than others while not sacrificing on the accuracy.
 
 \textit{Trick-GS} improves PSNR by 0.2dB on average while losing 50\% on the storage space and 15\% on the training time compared to \textit{Trick-GS-small}. The reduction on the efficiency with \textit{Trick-GS} is because of the use of progressive scale-based training and late densification that compensates for the loss from pruning of false positive Gaussians. 
 
 We tested an existing post-processing step~\cite{morgenstern2024compact} and further reduce the model size as low as $6MB$ and $12MB$ respectively for \textit{Trick-GS-small} and \textit{Trick-GS} over MipNeRF 360 dataset. The training time is not heavily impacted by the post-processing but the accuracy drop on $PSNR$ metric is $0.33dB$ which is undesirable for our method. Therefore, we leave the post-processing as a future work.

\input{figs/progressive_training.tex}

We compare \textit{Trick-GS} with two recent GS methods, Compact-GS~\cite{lee2024compact} and Mini-Splatting~\cite{fang2024minisplatting}. Tab.~\ref{tab:qual1} and Tab.~\ref{tab:qual_db} show the comparison of our approach over three datasets. Thanks to our trick choices, \textit{Trick-GS} learns models as low as $10MB$ for some outdoor scenes while keeping the training time around $10mins$ for most scenes. 

Learning to mask SH bands helps our approach to lower the storage space requirements. \textit{Trick-GS} lowers the storage requirements for each scene over three datasets even though it might result in more Gaussians than Mini-Splatting for some scenes. Our method improves some accuracy metrics over 3DGS while the accuracy changes are negligible. Advantage of our method is the requirement of $23\times$ less storage and $1.7\times$ less training time compared to 3DGS. 

Our approach achieves this performance without increasing the maximum number of Gaussians as high as the compared methods~\cite{fang2024minisplatting,lee2024compact}. Fig.~\ref{fig:num-gaussians} shows the change in number of Gaussians during training and an analysis on the number of pruned Gaussians based on the learned masks. \textit{Trick-GS} achieves comparable accuracy level while using $4.5\times$ less Gaussians compared to Mini-Splatting and $2\times$ less Gaussians compared to Compact-GS, which is important for the maximum GPU consumption on end devices. Fig.~\ref{fig:progressive-impact} shows the impact of our progressive training strategies. \textit{Trick-GS} obtains structurally more consistent reconstructions of tree branches thanks to the progressive training. 

\subsection{Ablation}

We evaluate the contribution of tricks in Tab.~\ref{tab:ablation} on \textit{MipNeRF360 - bicycle}  scene. Our tricks mutually benefits from each other to enable on-device learning. While Gaussian blurring helps to prune almost half of the Gaussians compared to 3DGS with a negligible accuracy loss, downsampling the image resolution helps to focus on the details by the progressive training and hence their mixture model lowers the training time and the Gaussian count by half. 

Significancy score based pruning strategy improves the storage space the most among other tricks while masking Gaussians strategy results in lower number of Gaussians at its peak and at the end of learning. Enabling progressive Gaussian scale based training also helps to improve the accuracy thanks to having higher number of Gaussians with the introduced split strategy. 

\input{figs/progressive_blur}

\begin{figure}[b!]
\centering

\subfloat{\includegraphics[width=0.8\linewidth]{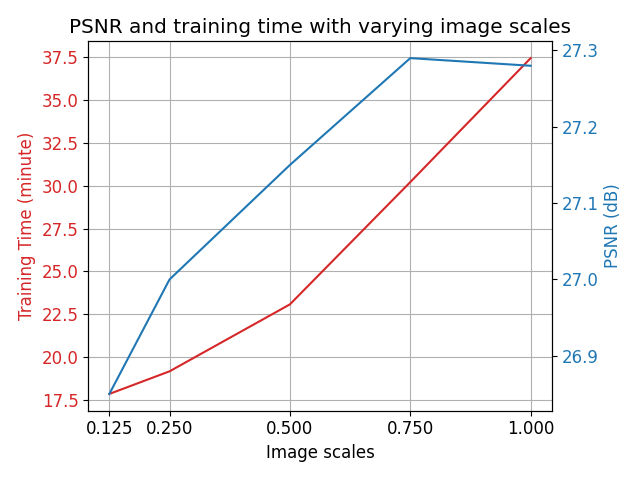}}
\caption{\small PSNR and training time evaluations~\wrt the lowest scale used to start a progressive resolution-based training.}
\label{fig:progressive_blur_plt}
\end{figure} 
We further evaluate, Fig.~\ref{fig:progressive_blur_plt}, the impact of changing image resolutions progressively and show that the improvement on the training time is much more than the loss on PSNR metric when when the training is started with a much smaller image resolution than the original resolution. Fig.~\ref{fig:progressive_blur} shows a visual example of changing the resolution progressively helping to recover tiny structures in the scene.

%% file: tables/dp_table.tex
\begin{table}[t!]
\centering
\resizebox{1.\linewidth}{!}{
\begin{tabular}{l|cccT{0.07\textwidth}T{0.05\textwidth}cT{0.08\textwidth}T{0.06\textwidth}}
\toprule
\rowcolor{lightgray} Dataset      & \multicolumn{8}{c}{Deep Blending}\\
\\\hline
Method & PSNR ↑ & SSIM ↑ & LPIPS ↓ & Storage ↓ (MB) & Time ↓ (min) &
FPS ↑ & Max-\#GS ↓ $\times1000$ & \#GS ↓ $\times1000$ \\
\toprule
\rowcolor{lightgray} \textsuperscript{*}3DGS~\cite{kerbl3Dgaussians}  & 29.41 & 0.903 & 0.243 & 676 & 36.03 & 137 & - & -\tabularnewline
\textsuperscript{*}INGP-base~\cite{muller2022instant} & 23.62 & 0.797 & 0.423 & 13 & 6.52 & 3 & - & -\tabularnewline
\rowcolor{lightgray} \textsuperscript{*}INGP-big~\cite{muller2022instant} & 24.96 & 0.817 & 0.390 & 48 & 8.00 & 3 & - & -\tabularnewline
\textsuperscript{*}Mini-Splatting~\cite{fang2024minisplatting} & 29.98 & 0.908 & 0.253 & - & - & - & - & 350\tabularnewline
\rowcolor{lightgray} \textsuperscript{*}Compact-GS~\cite{lee2024compact} & 29.79 & 0.901 & 0.258 & 43 & 27.55 & 181 & - & -\tabularnewline

\vspace{-2.5mm}
\\\hline
\bottomrule
\rowcolor{lightgray} 3DGS~\cite{kerbl3Dgaussians} & \thirdcell{29.46} & \thirdcell{0.899} & 0.266 & 664 & 25.29 & 121 & 2808 & 2808 \tabularnewline

Mini-Splatting~\cite{fang2024minisplatting} & \firstcell{29.78} & \firstcell{0.906} & \firstcell{0.251} & 80 & \thirdcell{16.53} & \firstcell{367} & 4521 & \firstcell{339}\tabularnewline

\rowcolor{lightgray} Compact-GS~\cite{lee2024compact} & \firstcell{29.78} & \secondcell{0.901} & \secondcell{0.258} & \thirdcell{41} & 37.59 & 182 & \thirdcell{2273} & 1043\tabularnewline

\vspace{-2.5mm}
\\\hline
\bottomrule
Trick-GS-small & \secondcell{29.63}	& \thirdcell{0.899}	& 0.266	& \firstcell{15}	& \firstcell{11.67}	& \secondcell{287} & \firstcell{1014} & \secondcell{363}\tabularnewline

\rowcolor{lightgray} Trick-GS & \thirdcell{29.46} & \thirdcell{0.899} & \thirdcell{0.260} & \secondcell{25} & \secondcell{13.11} & \thirdcell{260} & \secondcell{1308} & \secondcell{639}
\vspace{-1.5em}
\\
\bottomrule
\end{tabular}
}
\vspace{-0.3em}
\caption{Quantitative evaluation on Deep Blending dataset. Results with marked $^*$ method names are taken from the corresponding papers. Results below the double horizontal line are trained and evaluated on our system. }
\vspace{-1.5em}
\label{tab:qual_db}
\end{table}

%% file: tables/ablation.tex
\newcommand{\colwidth}{0.4cm}

\begin{table}[t!]
\centering
\resizebox{1.\linewidth}{!}{

\begin{tabular}{P{\colwidth}P{\colwidth}P{\colwidth}P{0.5cm}P{\colwidth}P{0.7cm}P{\colwidth}P{\colwidth}|P{0.9cm}P{0.9cm}P{0.9cm}P{1.1cm}P{0.8cm}P{0.7cm}P{1.3cm}P{1cm}}
\toprule
\rowcolor{lightgray} \multicolumn{8}{c}{Trick} & \multicolumn{8}{c}{MipNeRF 360 - Bicycle}\\ 
\hline
{\footnotesize BL}&{\footnotesize DS}&{\footnotesize SG}&{\footnotesize GM}&{\footnotesize AT}&\footnotesize{ SHM}&{\footnotesize DE}&{\footnotesize SC}& PSNR↑ & SSIM↑ & LPIPS↓ & Storage↓ (MB) & Time↓ (min) & FPS↑ & Max-\#GS↓ $\times1000$ & \#GS↓ $\times1000$\\
\hline
\rowcolor{lightgray} \xmark & \xmark & \xmark & \xmark & \xmark & \xmark & \xmark & \xmark & \secondcell{25.18} & \secondcell{0.76} & \firstcell{0.21} & 1409 & 33.36 & 70 & 5958 & 5958 \tabularnewline
\cmark & \xmark & \xmark & \xmark & \xmark & \xmark & \xmark & \xmark & 24.96 &	0.73 &	0.28 &	382 &	24.46 & 85 & 3398 &	3398 \tabularnewline
\rowcolor{lightgray} \xmark & \cmark & \xmark & \xmark & \xmark & \xmark & \xmark & \xmark & \firstcell{25.23} &	\firstcell{0.77} &	\firstcell{0.21} &	596 &	29.47 &	62 &	5294 &	5294 \tabularnewline
\xmark & \xmark & \cmark & \xmark & \xmark & \xmark & \xmark & \xmark & \thirdcell{25.15} &	\secondcell{0.76} &	\secondcell{0.22} &	233 &	25.54 &	132 & 4789 &	2067 \tabularnewline
\rowcolor{lightgray} \xmark & \xmark & \xmark & \cmark & \xmark & \xmark & \xmark & \xmark & 24.80 &	0.74 &	0.25 &	273 &	23.90 &	103 & 3462 &	2431 \tabularnewline
\xmark & \xmark & \xmark & \xmark & \cmark & \xmark & \xmark & \xmark & 25.15 &	\secondcell{0.76} &	\firstcell{0.21} &	677 &	29.7 &	86 &	6018 &	6018 \tabularnewline
\rowcolor{lightgray} \cmark & \cmark & \xmark & \xmark & \xmark & \xmark & \xmark & \xmark & 24.89 & 0.73 & 0.28 & 338 & 18.60 & 102 & 3002 & 3002 \tabularnewline
\cmark & \cmark & \cmark & \xmark & \xmark & \xmark & \xmark & \xmark & 24.84 &	0.72 &	0.29 &	128 &	\secondcell{14.91} &	173 &	2638 &	\thirdcell{1138} \tabularnewline
\rowcolor{lightgray} \cmark & \cmark & \cmark & \cmark & \xmark & \xmark & \xmark & \xmark &  24.62	& 0.70	& 0.31	& 75	& \firstcell{11.78}	& \secondcell{232}	& \firstcell{1838} &	\firstcell{670} \tabularnewline
\cmark & \cmark & \cmark & \cmark & \xmark & \cmark & \xmark & \xmark & 24.72 &	0.71 &	0.31 &	\firstcell{35} &	17.98 &	\thirdcell{223} &	1931 &	\secondcell{725} \tabularnewline
\rowcolor{lightgray} \cmark & \cmark & \cmark & \cmark & \cmark & \cmark & \xmark & \xmark & 24.75 & 0.71	& 0.31	& \secondcell{35} & \thirdcell{15.88} & \firstcell{262} & \thirdcell{1931} & \secondcell{725} \tabularnewline
\cmark & \cmark & \cmark & \cmark & \cmark & \cmark & \cmark & \xmark & 24.85 &	0.72 &	0.29 &	61 &	18.08 &	\thirdcell{223} & \secondcell{1920} &	1356 \tabularnewline
\rowcolor{lightgray} \cmark & \cmark & \cmark & \cmark & \cmark & \cmark & \cmark & \cmark & 25.07 &	\thirdcell{0.75} &	\thirdcell{0.25} &	\thirdcell{73} &	20.80 &	212 &	2374 &	1574
\\\bottomrule
\end{tabular}}
\caption{\small Ablation study on tricks adopted by our approach using `bicycle' scene. Our tricks are abbreviated as BL: progressive Gaussian blurring, DS: progressive downsampling, SG: significance pruning, GM: Gaussian masking, SHM: SH masking, AT: accelerated training, DE: late densification, SC: progressive scaling. Our full model \textit{Trick-GS} uses all the tricks while \textit{Trick-GS-small} uses all but DE and SC.}
\label{tab:ablation}
\end{table}

%% file: figs/num_gaussians.tex
\begin{figure}[t!]
\centering
\captionsetup[subfigure]{labelformat=empty}
\subfloat[]{\includegraphics[width=0.8\linewidth]{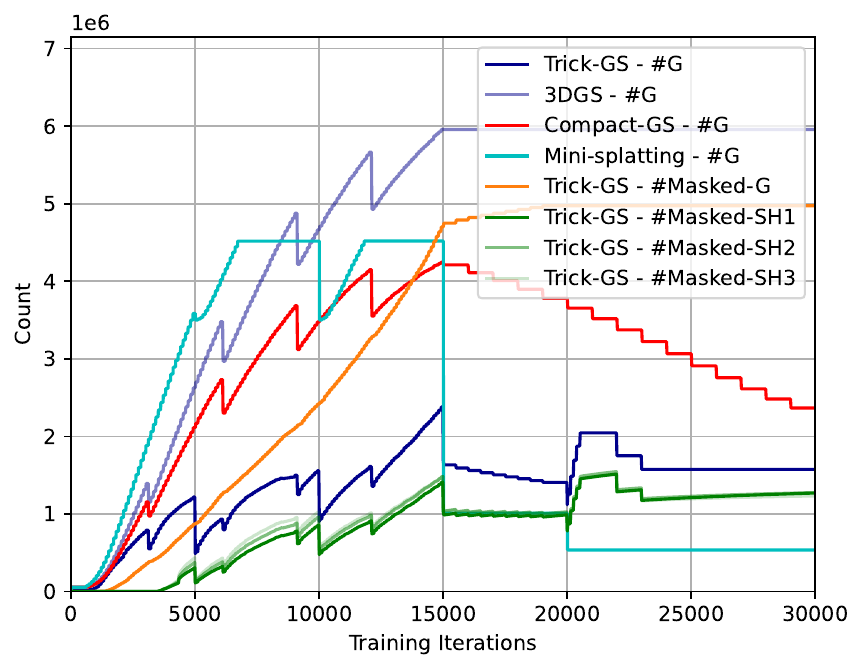}}
\caption{\small Number of Gaussians (\#G) during training (on MipNeRF 360 - \textit{bicycle} scene) for all methods, number of masked Gaussians (\#Masked-G) and number of Gaussians with a masked SH band for our method. Our method performs a balanced reconstruction in terms of training efficiency by not letting the number of Gaussians increase drastically as other methods during training, which is a desirable property for end devices with low memory.}
\label{fig:num-gaussians}
\end{figure} 

%% file: figs/progressive_training.tex
\begin{figure}[b!]
\centering
\captionsetup[subfigure]{labelformat=empty}
\subfloat[GT]{\includegraphics[width=0.49\linewidth]{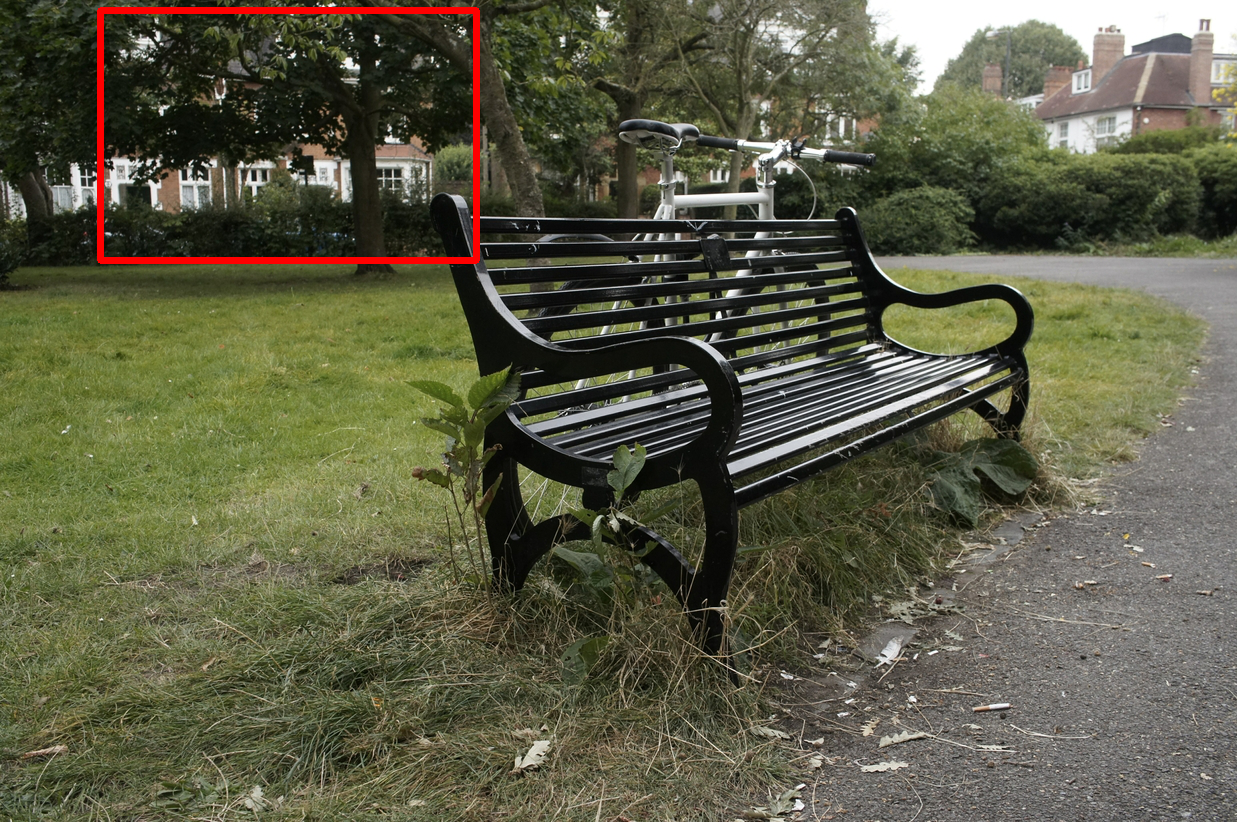}} 
\subfloat[Predictions]{\includegraphics[width=0.49\linewidth]{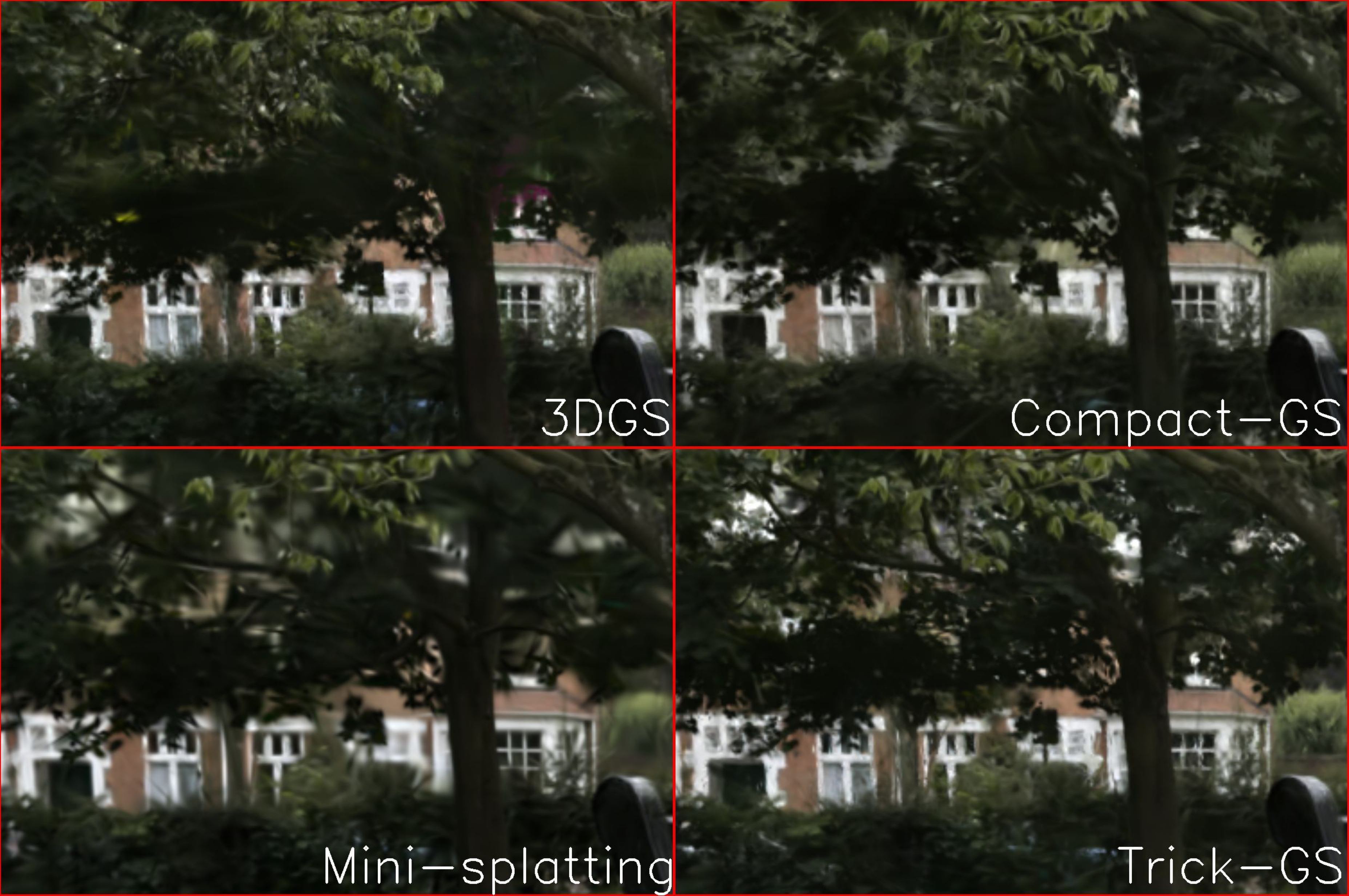}} 
\caption{Impact of progressive training strategies on challenging background reconstructions. We empirically found that progressive training strategies as downsampling, adding Gaussian noise and changing the scale of learned Gaussians have a significant impact on the background objects with holes such as tree branches. }
\label{fig:progressive-impact}
\end{figure}

%% file: figs/progressive_blur.tex
\begin{figure}[t!]
\centering
\captionsetup[subfigure]{labelformat=empty}
\subfloat[vanilla]{\small 3DGS} \hspace{6.5mm}\raisebox{-.37\height}{
\subfloat[]{\includegraphics[width=0.25\linewidth, height=0.25\linewidth]{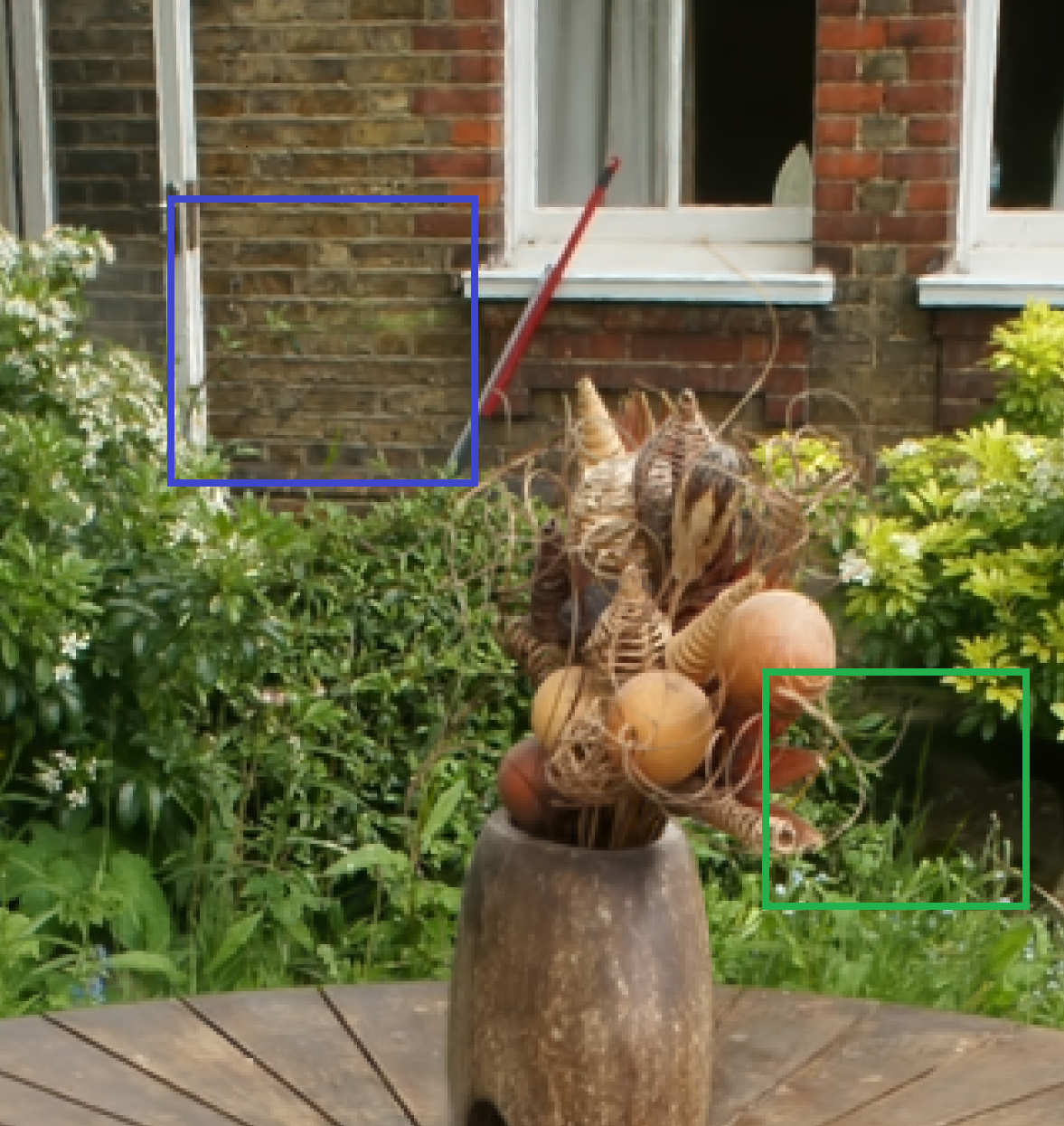}}
\subfloat[]{\includegraphics[width=0.25\linewidth, height=0.25\linewidth]{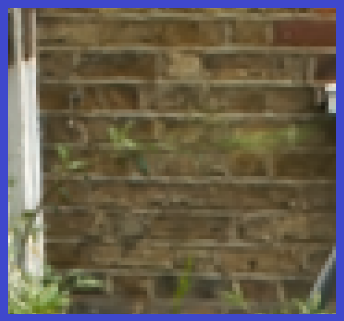}}
\subfloat[]{\includegraphics[width=0.25\linewidth, height=0.25\linewidth]{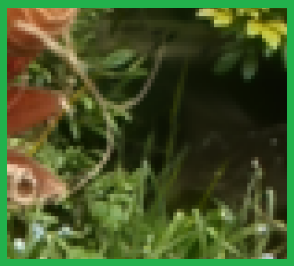}}} \\
\vspace{-8mm}

\subfloat[resolution\\change]{\small Progressive}  \raisebox{-.57\height}{
\subfloat[(a) Prediction]{\includegraphics[width=0.25\linewidth, height=0.25\linewidth]{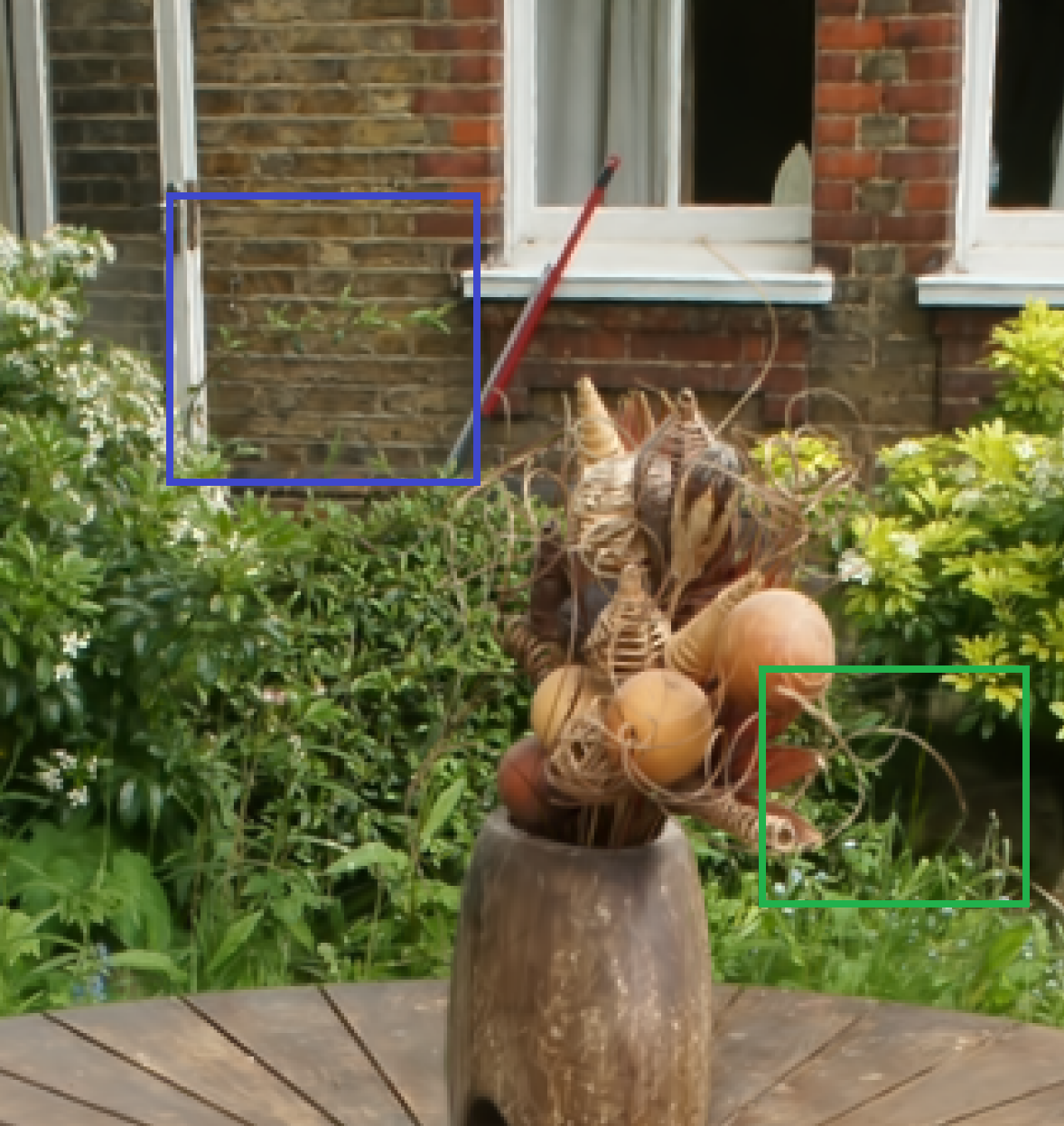}}
\subfloat[(b) Improvement 1]{\includegraphics[width=0.25\linewidth, height=0.25\linewidth]{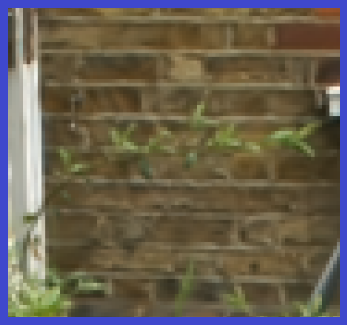}}
\subfloat[(c) Improvement 2]{\includegraphics[width=0.25\linewidth, height=0.25\linewidth]{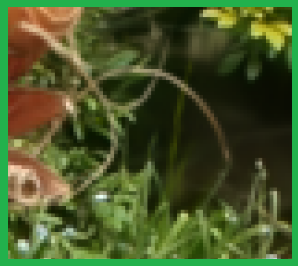}}\vspace{-3mm}}
\caption{\small Visual results (a) from vanilla 3DGS (first row) and a model trained with progressive resolution based strategy (second row) starting with scale $0.125$. We use `garden` from MipNeRF360 dataset and zoom into the improvements (b) \& (c) for clarity.}
\label{fig:progressive_blur}
\end{figure}

%% file: conclusion.tex
\section{Conclusion}
\label{conclusion}
We have proposed a mixture of strategies adopted from the literature to obtain compact 3DGS representations. We have carefully designed and chosen strategies from the literature and showed competitive experimental results. Our approach reduces the training time for 3DGS by $1.7\times$, the storage requirement by $23\times$, increases the FPS by $2\times$ while keeping the quality competitive with the baselines. The advantage of our is method being easily tunable \wrt~the application/device needs and it further can be improved with a post-processing stage from the literature \eg~codebook learning, Huffman encoding. We believe a dynamic and compact learning system is needed based on device requirements and therefore leave automatizing such systems for future work.